\pdfoutput=1

\documentclass[11pt]{article}

\usepackage{tabularx} 
\usepackage[table]{xcolor}
\usepackage{acl}

\usepackage{times}
\usepackage{latexsym}
\usepackage{booktabs}
\usepackage{xcolor}
\definecolor{mydarkgreen}{rgb}{0.0, 0.5, 0.0}
\newcommand{\mydarkgreen}[1]{{\color{mydarkgreen}#1}}
\usepackage{amssymb}
\usepackage[T1]{fontenc}

\usepackage[utf8]{inputenc}
\usepackage{algorithm}
\usepackage{microtype}
\usepackage{makecell}
\usepackage{pifont}
\usepackage{arydshln}
\usepackage{multirow}
\usepackage{ulem}
\usepackage{multicol}
\usepackage{algpseudocode}  
\usepackage{amsmath,amssymb,amsfonts}
\usepackage{subfigure}
\usepackage{marvosym}
\usepackage{amsthm}
\usepackage{pgfplots}
\definecolor{mygray}{gray}{0.8}
\usepackage{tabularx}
\pgfplotsset{compat=1.16}
\usepackage{newfloat}
\usepackage{listings}
\usepackage{CJK}
\usepackage{microtype}

\usepackage{inconsolata}

\usepackage{color}

\usepackage{colortbl}
\definecolor{mygray}{gray}{0.8}

\newcommand{\black}[1]{{\color{black}{#1}}}

\newcommand{\myred}{\black}
\title{Piecing Together Clues: A Benchmark for Evaluating the Detective Skills of Large Language Models}

\author{Zhouhong Gu\textsuperscript{\rm $\spadesuit$},
Lin Zhang\textsuperscript{\rm $\spadesuit$},
Jiangjie Chen\textsuperscript{\rm $\spadesuit$},
Haoning Ye\textsuperscript{\rm $\spadesuit$},
Xiaoxuan Zhu\textsuperscript{\rm $\spadesuit$},\\
\bf
Zihan Li\textsuperscript{\rm $\spadesuit$},
Zheyu Ye\textsuperscript{\rm $\heartsuit$},
Yan Gao\textsuperscript{\rm $\heartsuit$},
Yao Hu\textsuperscript{\rm $\heartsuit$},
Yanghua Xiao\textsuperscript{\rm $\spadesuit$}\thanks{Corresponding authors.},
Hongwei Feng\textsuperscript{\rm $\spadesuit$}\footnotemark[1]
\\
\textsuperscript{\rm $\spadesuit$}Fudan University\quad
\textsuperscript{\rm $\heartsuit$}Xiaohongshu Inc.\\
\texttt{\{zhgu22, linzhang22, xxzhu22, zhli21\}@m.fudan.edu.cn},\\
\texttt{\{zheyuye, yadun, xiahou\}@xiaohongshu.com},\\
\texttt{\{jjchen19, shawyh,jjchen19\}@fudan.edu.cn}}

\begin{document}
\begin{CJK}{UTF8}{gbsn}
\maketitle

\begin{abstract}

Detectives frequently engage in information detection and reasoning simultaneously when making decisions across various cases, especially when confronted with a vast amount of information. With the rapid development of large language models~(LLMs), evaluating how these models identify key information and reason to solve questions becomes increasingly relevant. We introduces the DetectBench, a reading comprehension dataset designed to assess a model's ability to jointly ability in key information detection and multi-hop reasoning when facing complex and implicit information. The DetectBench comprises 3,928 questions, each paired with a paragraph averaging 190 tokens in length. To enhance model's detective skills, we propose the Detective Thinking Framework. These methods encourage models to identify all possible clues within the context before reasoning. Our experiments reveal that existing models perform poorly in both information detection and multi-hop reasoning. However, the Detective Thinking Framework approach alleviates this issue.

\end{abstract}

\section{Introduction}
\begin{figure}[t]
    \centering
    \resizebox{0.8\columnwidth}{!}{
    \includegraphics{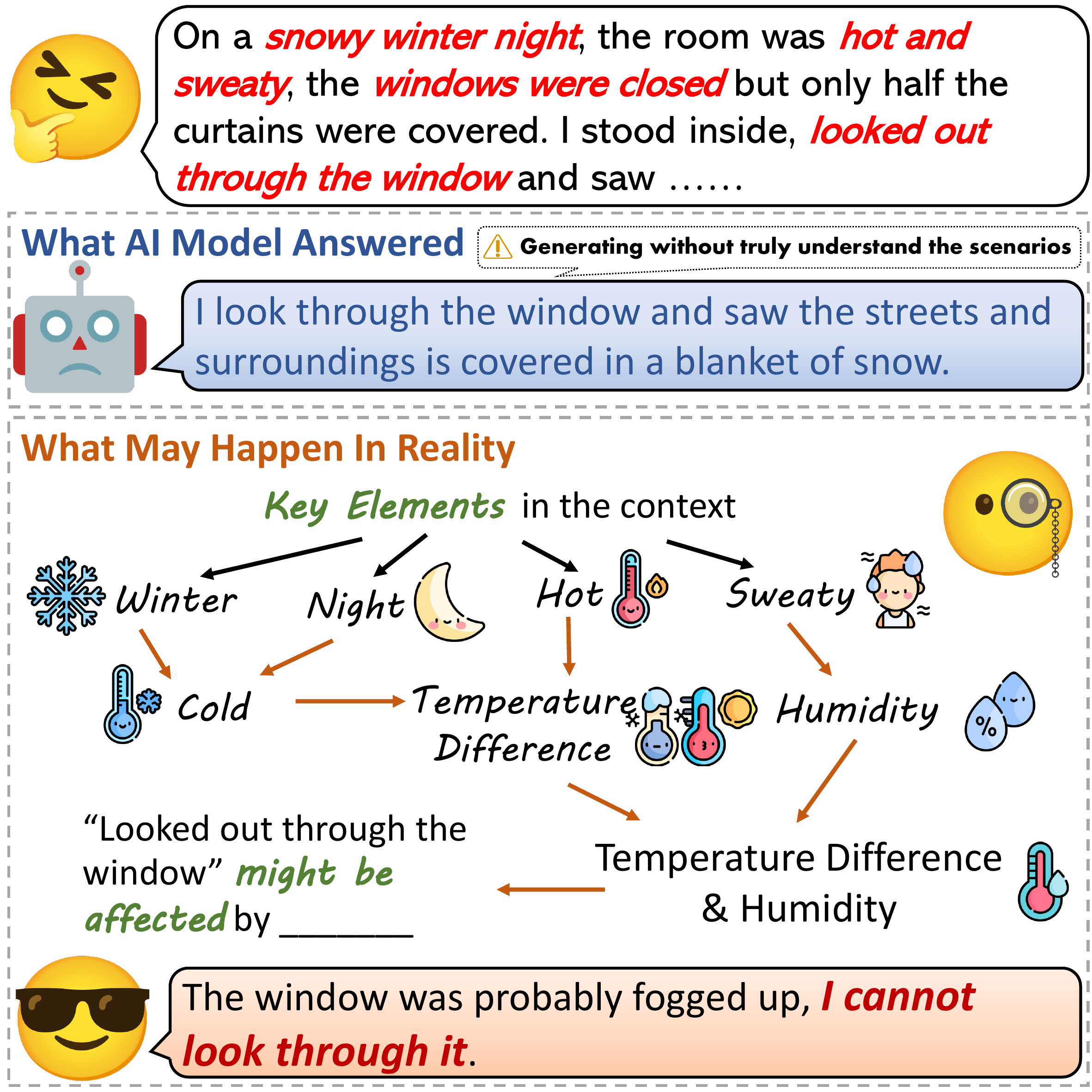}
    }
    \caption{
    When facing overloaded information LLMs may produce outputs arbitrarily due to their inability to engage in deep contemplation.
    In contrast, humans who are experienced, like detectives, analyze and correlate all available information, thereby identifying pivotal clues that lead to the answer of the problem.    
    }
    \label{fig:intro}
    \vspace{-5mm}
\end{figure}

When handling vast amounts of information across various cases, experienced detectives typically engage in a continuous process of discovering clues, conducting reasoning, updating clues and updating reasoning.
This ability to combine clue detection and reasoning is a critical aspect of detective skills.
{It is only when the key clues are found that effective reasoning can be carried out to solve the problem.}
As the LLM evolves, user often inject complex, massive amounts of reference to it when solving problems.
This raises a question: \textit{when faced with overloaded information, does the LLM also possess detective skills to recognize clues and use it to reason effectively and solve problems?}

Clues, often concealed within seemingly unrelated and dispersed information, is seldom straightforward in real-cases. 
This necessitates models capable of discerning the essences and reasoning from them.
For example, as shown in Figure~\ref{fig:intro}, only when we realize that changes in temperature and humidity can make glass foggy, can we figure out that details about temperature and humidity are key to seeing through the glass.
There has many existing tasks evaluate the model's joint abilities in clues detection and reasoning, such as reading comprehension~\cite{yu2020reclor,kazi2021uquad1,lu2022contextual}, retrieval reasoning~\cite{yang2018hotpotqa,chen2023benchmarking}, and fact verification~\cite{thorne2018fever,Thorne19FEVER2,Aly21Feverous}.
However, for the existing benchmarks in these tasks, important information is clustered together and often linked directly to questions that can be found by keywords.

Drawing inspiration from the detective's process of detecting and reasoning from a vast array of implicit information, we propose a multiple-choice question answering benchmark called \textbf{DetectBench}.
This benchmark comprises 3,928 questions, each paired with a paragraph averaging 190 tokens.
The purpose of DetectBench is to evaluate the proficiency of models in identifying and reasoning from clues within a complex context to answer questions.
The characteristics of DetectBench include:
1. Clues related to problem-solving do not directly appear within the context.
2. It necessitates the combination of multiple clues within the context to derive more critical clues.
3. The context contains a significant amount of misleading and irrelevant information.

We collect a large number of detective puzzles from open-source platforms and rewrit them into DetectBench, which consists of context, question, options, answer, and an explanation of the answer.
The context in DetectBench mimics the intricate stories, situations, and character interactions found in detective puzzles.
Each context pair includes a question, options, and an answer, providing a challenging test of the model's ability to detect and reason with clues.
Each question is accompanied by an explanation of how to arrive at the answer, utilising ``Clue Graphs'' as shown at the bottom of Figure~\ref{fig:intro}.
These graphs begin with text that exactly matches the basic clues in context, and then gradually connect the clues into new ones, leading directly to the answer.

In experiments conducted on human participants and LLMs, we assessed their abilities to clue detection and accuracy in answer questions.
Our findings indicate that humans significantly outperformed the most advanced LLMs in both tasks, suggesting that most LLMs have poorer detective skills than the average person.
We also found that performance in clue detection correlates with performance in answering questions. 
This demonstrates the effectiveness of the annotation in DetectBench and emphasizes the importance of detecting clues before reasoning.

To enhance model's detective skills, we proposed a baseline method to jointly enhance model's clue detection ablities and reasoning abilities: Detective Thinking.
Just like experienced detective combine clues detection and reasoning, Detective Thinking enhance the detective skills of existing LLMs by guiding them to consider all possible clues comprehensively, reason, and then summarize all the reasoning process to fine clues, and finally reason from the clue to find the answer to the question.
Using the Detective Thinking improved model's performance in both clues detection and reasoning, but the performance is still greatly leg behind human participants.

In summary, the main contributions of this study include:
(1) The introduction of the DetectBench, providing a new standard for assessing model's key information detection and reasoning abilities.
(2) The development of the Detective Thinking Prompt and Detective Thinking Finetune method, significantly enhancing model's joint performance in information detection and reasoning abilities.
(3) Through extensive experiments, the limitations of existing models in discovering key information and conducting deep reasoning are verified, and it was shown that these limitations could be mitigated after using the Detective Thinking Prompt/Finetune methods.

\section{Related Works}

\begin{table*}[!ht]
\centering
\resizebox{\textwidth}{!}{
\begin{tabular}{lllcll}
\hline
Benchmark & \# of Questions & Ave. Length~\footnote{Here only count the token contain in the context that is related to answer the question.} & Explanation to Answer & Ansering Format & Metrics \\ \hline

HotpotQA~\cite{yang2018hotpotqa}    & 112,779   & 137.9 & & Free Text & Rouge     \\ \hline
FEVER~\cite{thorne2018fever}    &  185,445 & 9.4 & \mydarkgreen{$\checkmark$} &\makecell[l]{Classification\\\&Text retrieval} & Accuracy \& F1 \\ \hline
HellaSwag~\cite{zellers2019hellaswag}   & 59,950    & 38.5  & & Choice QA & Accuracy  \\ \hline
Reclor~\cite{yu2020reclor}      & 6,138     & 66.4  & & Choice QA & Accuracy     \\ \hline
WinoGrande~\cite{sakaguchi2021winogrande}  & 12,282    & 21.1  & \mydarkgreen{$\checkmark$}  & Choice QA & Accuracy  \\ \hline
QReCC~\cite{anantha2020open} &79,952&/&&Free text&\makecell[l]{Rouge \& Recall}\\ \hline
FEVEROUS~\cite{Aly21Feverous} &87,026&25.3&\mydarkgreen{$\checkmark$}&\makecell[l]{Classification\\\&Text retrieval}& Accuracy \& F1\\ \hline
TopiocQA~\cite{adlakha2022topiocqa} &50,466&145.2&&Free text&EM \& F1\\ \hline
INSCIT~\cite{wu2023inscit} &4,712&/&&Free text&BLEU \& F1\\ \hline
RGB~\cite{chen2023benchmarking} &100&26.3&\mydarkgreen{$\checkmark$}&Free text& EM\\ \hline
\multirow{2}{*}{DetectBench} & \multirow{1}{*}{396(train)+1928(de-} & \multirow{2}{*}{190.2} & \multirow{2}{*}{\mydarkgreen{$\checkmark$}}  & Choice QA \& Free &Accuracy \\
&v)+1604(test)=3,928&&&Text Reasoning& \& Rouge\\ \hline
\end{tabular}}
\caption{The comparison between the DetectBench with other and Information Retrieval Benchmarks and Common Sense Reasoning Benchmarks.}
\vspace{-5mm}
\end{table*}

\subsection{Information Retrieval}

The domain of Information Retrieval aims to address pertinent tasks through the extraction of crucial data from a plethora of references, where the most significant challenge lies in the identification of implicit key information~\cite{zhu2023large,yang2022survey}. Traditional benchmarks in Information Retrieval have historically segmented the task of Information Extraction for the purpose of evaluating models independently~\cite{martinez2020information,cheng2021hacred,lu2022unified}. Recent endeavors, however, have led to the development of benchmarks designed for the holistic assessment of task resolution capabilities. 
Among these, HotPotQA~\cite{yang2018hotpotqa} necessitates the discovery of question-relevant information across paragraphs to aid in response formulation, FEVER~\cite{thorne2018fever,Thorne19FEVER2,Aly21Feverous} necessitates the identification of evidentiary support to validate or negate a claim, and RECLOR~\cite{yu2020reclor}, UQuAD~\cite{kazi2021uquad1}, BIOMRC~\cite{lu2022contextual} emphasizes the extraction of text segments pivotal for answering queries.
Nonetheless, the linkage between key information and queries within these benchmarks is overtly conspicuous, allowing for the location of pertinent data through string matching techniques and facilitating correct answer derivation via one or two inferential leaps.

However, the unique feature of the DetectBench is its reliance on evidence that is widely dispersed and implicit to answer questions.

\subsection{Commonsense Reasoning}

\begin{table}[t]
\centering
\resizebox{\columnwidth}{!}{
\begin{tabular}{llll}
\hline
Type                        & Example                                       &\#                       &\% \\\hline
\multirow{3}{*}{How}       &\textit{``How was the murder weapon}          &\multirow{3}{*}{1,647}  &\multirow{3}{*}{41.9}\\
                            &\textit{handled such that it was not}&&\\
                            &\textit{discovered at the scene?''}&&\\
\hline
\multirow{2}{*}{What}       &\textit{``What's the house number}             &\multirow{2}{*}{731}    &\multirow{2}{*}{18.6}\\
                            &\textit{where Smith lives?''}&&\\
\hline
\multirow{3}{*}{Which}     &\textit{``Which building doesn't have}    &\multirow{3}{*}{498}    &\multirow{3}{*}{12.7}\\
                            &\textit{any graduatestudents living in}&&\\
                            &\textit{this dormitory building?''}&&\\
\hline
\multirow{2}{*}{Who}        &\textit{``Who is the murderer of the}              &\multirow{2}{*}{459}    &\multirow{2}{*}{11.7}\\
                            &\textit{painter?''}&&\\
\hline
\multirow{1}{*}{Why}        &\textit{``Why did Harry suspect Filch?''}      &\multirow{1}{*}{378}    &\multirow{1}{*}{9.6}\\
\hline
\multirow{1}{*}{When}       &\textit{``When is Teacher's birthday?''}       &\multirow{1}{*}{167}    &\multirow{1}{*}{4.3}\\
\hline
\multirow{2}{*}{Where}      &\textit{``Where exactly does woman}&\multirow{2}{*}{121}   &\multirow{2}{*}{3.1}\\
                            &\textit{come from?''}&&\\
\hline
\hline
\multirow{3}{*}{Other}     &\textit{``Please determine the respective}    &\multirow{3}{*}{378}    &\multirow{3}{*}{9.6}\\
                            &\textit{professions of Faulkner, Santiago,}&&\\
                            &\textit{and Hemingway.''}&&\\
\hline
\end{tabular}}
\caption{All eight question type in Detective Reasoning and their frequency.}
\label{tab:human_performance}
\vspace{-5mm}
\end{table}
The exploration of Commonsense Reasoning encompasses a variety of research efforts, traditionally classified into single-hop reasoning, multi-hop reasoning, and reasoning that is uncommon yet plausible.
Datasets facilitating single-hop reasoning, such as HellaSwag~\cite{zellers2019hellaswag} and WinoGrande~\cite{sakaguchi2021winogrande}, present challenges in reasoning through narrative continuation, where the difficulty often resides in the formulation of options and potentially in the design of adversarial options aimed at undermining specific models.
Multi-hop reasoning benchmarks like StrategyQA~\cite{geva2021did} annotate the reasoning path, concentrating on the capacity of models to execute multi-hop reasoning in response to questions.
Reasoning that is uncommon yet feasible, as demonstrated in datasets like α-NLG~\cite{bhagavatula2019aNLG}, d-NLI~\cite{rudinger2020bNLG}, and UnCommonsense Reasoning~\cite{zhao2023uncommonsense, arnaout2022uncommonsense}, typically originates from pre-existing datasets by selecting the least likely option as the correct response and elucidating the rationale behind this selection.

The DetectBench is categorized as uncommon but plausible multi-step reasoning, feature on finding where to start such reasoning tasks.
The process of reasoning usually starts with small details that might seem unimportant.
But, when looked at more closely, these details help show a clear path that leads to a clear answer.


\section{Benchmark Construction}

\begin{figure*}[t]
    \centering
    \resizebox{0.85\textwidth}{!}{
    \includegraphics{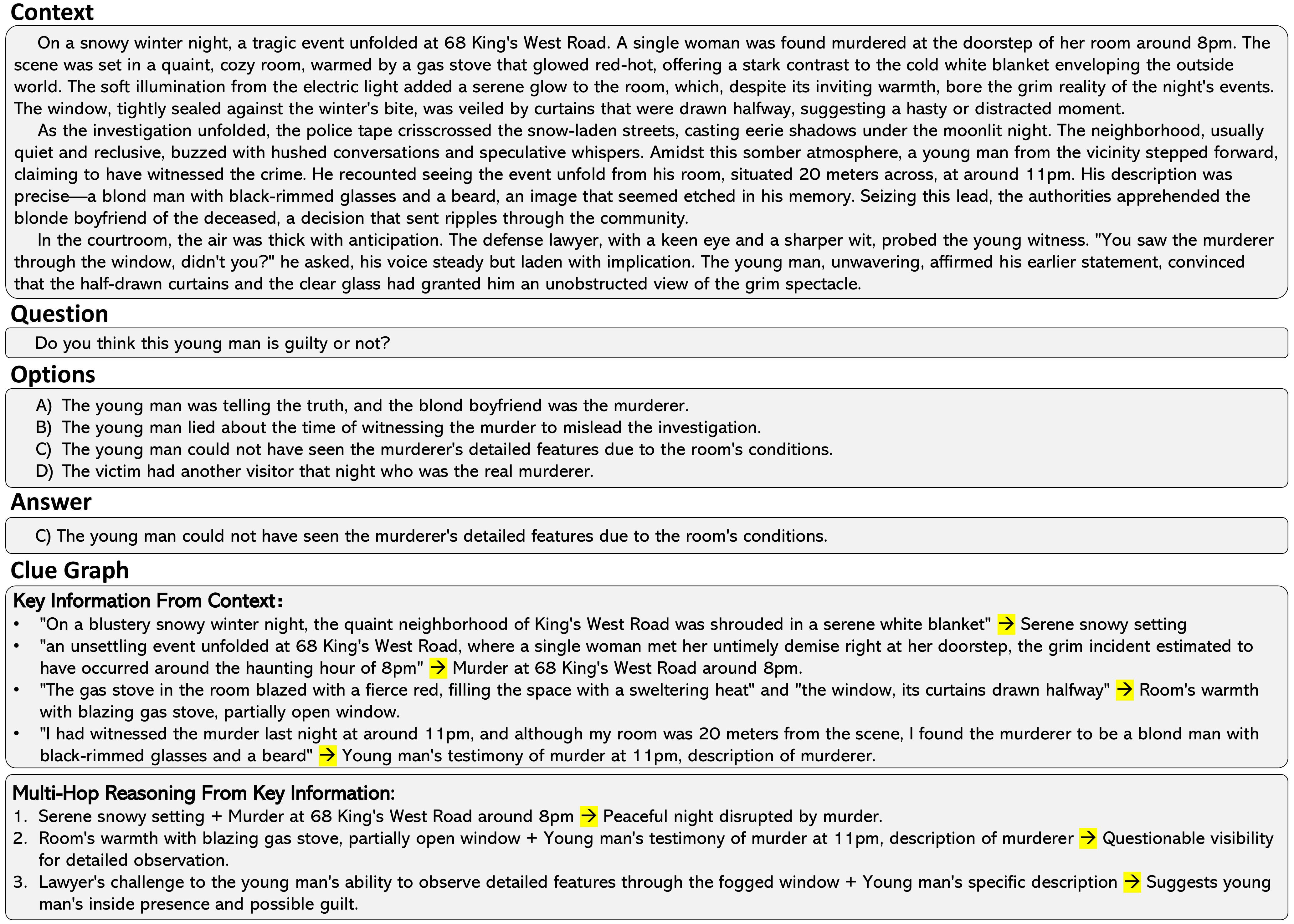}
    }
    \caption{The example of the question in DetectBench}
    \label{fig:example}
    \vspace{-5mm}
\end{figure*}

\begin{figure*}[t]
    \centering
    \resizebox{0.9\textwidth}{!}{
    \includegraphics{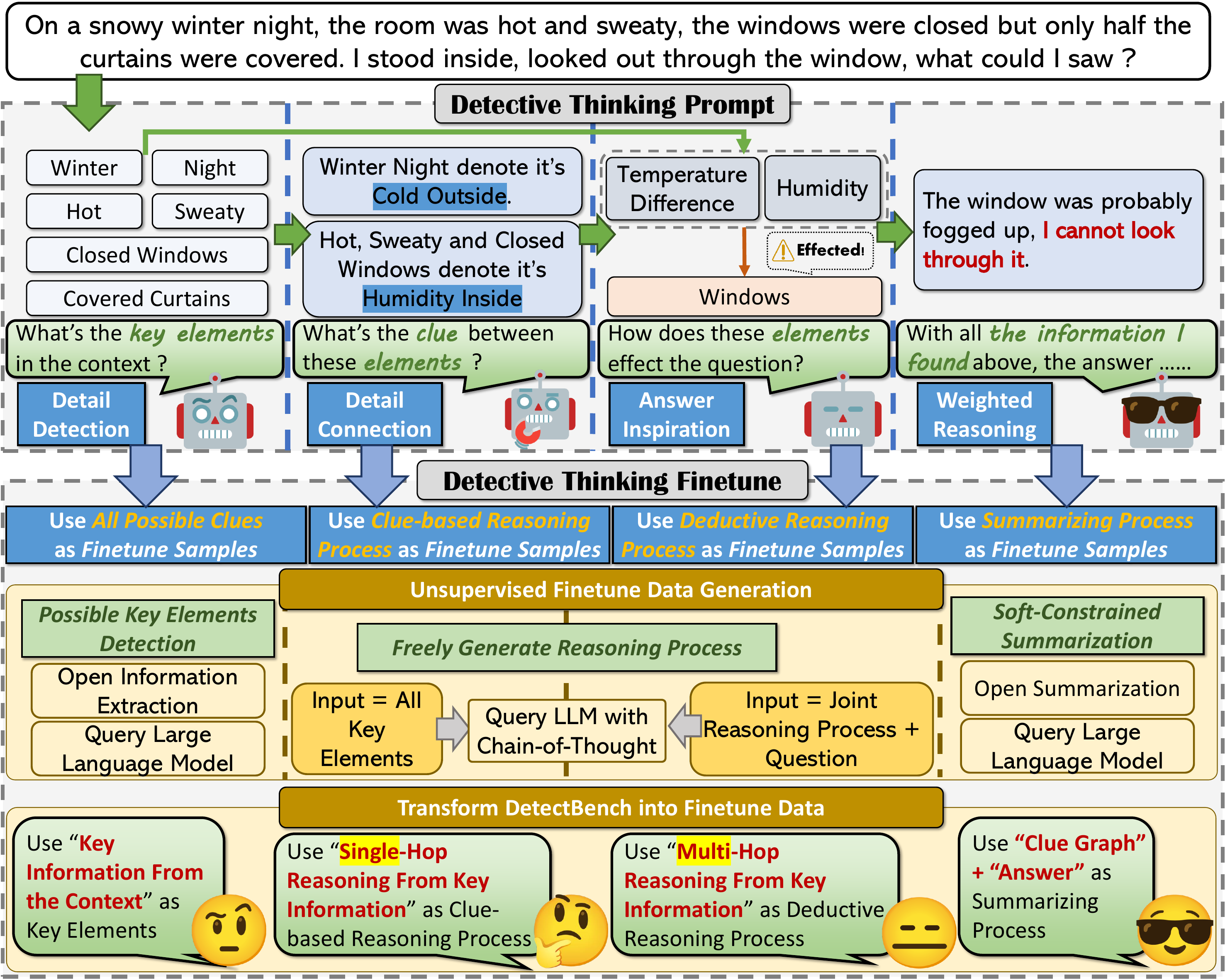}
    }
    \caption{
    Within the Detective Thinking Prompt paradigm, the process is bifurcated into distinct phases: Detail Detection and Detail Connection, followed by Answer Inspiration and Weighted Reasoning. 
    The Detective Thinking Finetune strategy is predominantly aimed at collecting data for fine-tuning.
    The first three phases permits free generation via open-source models, culminating in the aggregation of these outputs into a cohesive answer during the final stage.
    }
    \label{fig:framework}
    \vspace{-5mm}
\end{figure*}

\begin{table}[t]
\centering

\fontsize{8pt}{10pt}\selectfont
\begin{tabular}{l|l}
\hline
 \multicolumn{2}{l}{Human Performance}\\
\hline
Average Accuracy & 74.1\% \\
Top Accuracy & 93.3\% \\
Lowest Accuracy & 53.3\% \\
\hline
\end{tabular}
\caption{
Human performance in answering questions.}
\label{tab:human_experiments}
\vspace{-5mm}
\end{table}

\subsection{Benchmark Construction}
The DetectBench aims to evaluate model's detective skills, which means joint abilities in clues detection and multi-step commonsense reasoning.
Therefore, benchmark should provide the following elements:
(1). Question should not contain ethical integrity or encompass topics of a sensitive nature.
(2). Question descriptions should contain lengthy, complex, seemingly unrelated and even mislead information.
(3). The solution to the question should involve multi-step reasoning based on the original information.
(4). The model's response to the question should be assessed objectively and accurately.

Each question in DetectBench is organized in JSON format, comprising five main elements: ``Context'', ``Question'', ``Options'', ``Answer'' and ``Clue Graph'' as shown in Fig.~\ref{fig:example}.
Data processing includes question selection, question rewriting, and manual verification stages, with the first two stages primarily assisted by the GPT-4-turbo-1106-preview model.

\textbf{Question Selection:}
To ensure the benchmark focuses on ``clues detection'' and ``multi-step commonsense reasoning'', we carry out a detailed verification on all questions.
Given the potential for multiple answers and reasoning processs in detective reasoning questions, we endeavored to ensure each question's reasoning process is as clear and straightforward as possible to ensure the reasonableness and uniqueness of the answer and reasoning process.
Simultaneously, we excluded questions overly rely on symbolic logic or specialized knowledge because such questions cannot be solved simply by retrieving related information or clues but also domain knowledge and special training technique.
Specifically, we excluded five types of questions:
1. Questions that are not ethical or have sensitive matters.
2. Questions requiring visual or auditory information for support;
3. Questions that are anti-logical, have unreasonable answers, or are overly diverse;
4. Questions requiring extensive symbolic logic or domain knowledge;
5. Questions with overly obvious key information.

\textbf{Question Rewriting:}
The questions in the original text may mix the problem description with the question, sometimes even provide the answer directly or lack relevant information to do reasoning.
Therefore, we first use \underline{``Context''} and \underline{``Question''} to distinguish between the background and the query of the question.
Then the original natural text questions are converted into a multiple-choice format. 
The converted format includes \underline{``Options''} and \underline{``Answer''} fields to represent the choices and the correct answer.
Additionally, we constructed a \underline{``Clue Graph''} to explicitly represent the reasoning process. 
We annotated important content within the context as \underline{``Key Information from Context''}. 
Based on this key information, we delineated the \underline{``Multi-Hop Reasoning From Key Information''}, which encompasses the reasoning process from a single piece of information as well as joint reasoning based on multiple pieces of information. 

\textbf{Manual Verification:}
All questions processed by the GPT-4-turbo-1106-preview model undergo manual verification.
Five annotators are recruited to work with the authors on verification.
This includes initial screening to eliminate questions with unreasonable answers or options that require significant modification.
Additionally, detail adjustment is performed to fine-tune options and answers to make them more reasonable and natural.
The Appendix~\ref{appendix_annotators} provides detailed requirements and examples for annotation.


\subsection{Human Performance}

To examine human detective skills, we invited 50 participants to answer questions from the DetectBench dev set.
The examination took a total of three hours, and participants were allowed to leave early if they completed the task.
The participants were comprised of undergraduate and graduate students from universities across China, each remunerated at rates exceeding the local minimum hourly wage and bonuses for each correctly answered question. 
To facilitate human participation, we translate the benchmark into Chinese and used an online question-and-answer platform to collect answers and measure time spent.
Each participant was asked to answer 15 questions from a subset of 250 questions from the DetectBench development set, which ensured that each question was answered by three different participants.
The results of human performance on the DetectBench are presented in Table~\ref{tab:human_performance}.



\section{Detective Thinking Method}
\subsection{Detective Thinking Prompt}
The Detective Thinking Prompt is intended to help the model identify crucial information and extract precise answers through progressively deeper logical reasoning, as demonstrated in Fig.~\ref{fig:framework}.
Specially, Detective Thinking Prompt consists of four stage:
\textbf{(1) Detail Detection}, which aims to prompt the model to uncover details and facts within the given content, particularly those not explicitly stated in the original text.  
\textbf{(2) Detail Association}, which requires the model to comprehend the inherent connections between pieces of information in the text and to generate new related information based on identified details.
\textbf{(3) Answer Inspiration} involves identifying key information necessary for solving a question and initiating reasoning around this information to trigger possible answers.
\textbf{(4) Weighted Reasoning} reinforces the model's reliance on its generated reasoning outcomes in the determination of the final answer compared to the overall context.
For detailed prompts for each stage, please refer to Appendix~\ref{appendix_prompt}.

\subsection{Detective Thinking Finetune}

Building upon the aforementioned Detective Thinking Prompt, we propose a finetuning strategy to further improve model's detective skills, as illustrated in the below of Fig.~\ref{fig:framework}.
For benchmarks that have reasoning processes explicitly annotated, such as DetectBench, one can concatenate the reasoning outputs for each stage in the Detective Thinking Prompt as the finetuning data. 
For benchmarks that have only standard answers, the model can automatically complete the reasoning process based on the questions and answers, and then organise this reasoning content as fine-tuning data.
This method has the advantage of using the freely output LLM as fine-tuning data in the first three stages. 
This significantly reduces the complexity of constructing datasets containing inference processes.

\section{Experiments}

\begin{table*}[ht]
    \centering
    \resizebox{0.9\textwidth}{!}{
    \begin{tabular}{|l|cc|cc|cc|cc|cc|cc|cc|cc|cc|}
    \hline
         &
         \multicolumn{2}{c|}{GPT4} & \multicolumn{2}{c|}{GPT35} & \multicolumn{2}{c|}{GLM4} & \multicolumn{2}{c|}{ChatGLM3-chat} & \multicolumn{2}{c|}{ChatGLM3-base} & \multicolumn{2}{c|}{Llama2-chat} & \multicolumn{2}{c|}{Llama2-base} \\
        &KeyInfo.&Acc.&KeyInfo.&Acc.&KeyInfo.&Acc.&KeyInfo.&Acc.&KeyInfo.&Acc.&KeyInfo.&Acc.&KeyInfo.&Acc.\\
        \hline
        
        \multicolumn{15}{|l|}{\cellcolor{mygray} \textit{Naive Questioning}} \\
        \hline
        Naive                           & 44.4 & 56.5 & 15.3 & 33.0 & 31.1 & 40.2 & 15.3 & 41.3 & 9.71 & 39.6 & 10.8 & 47.5 & 10.7 & 39.6 \\
        Naive (3-shot)                  & 40.6 & 54.4 & 15.3 & 34.9 & 30.3 & 39.4 & 10.8 & 41.8 & 13.1 & 42.3 & 11.5 & 47.1 & 9.9 & \textbf{41.4} \\

        \hline
        \multicolumn{15}{|l|}{\cellcolor{mygray} \textit{Process Enhanced Method}} \\
        \hline
        
        Self-CoT                        & 31.4 & 60.7 & 17.73 & 32.3 & 31.0 & 45.1 & 17.0 & 40.4 & 21.8 & 35.4 & 20.6 & 50.6 & 16.6 & 38.7 \\
        Auto-CoT (3-shot)               & 37.5 & 56.7 & 19.91 & 33.9 & \textbf{35.5} & 43.2 & 18.1 & 41.3 & 22.9 & 37.5 & 20.4 & 47.5 & 19.9 & 40.9 \\

        \hline
        \multicolumn{15}{|l|}{\cellcolor{mygray} \textit{Output Ensemble Method}} \\
        \hline
        
        Self-Consistency                & 31.7 & 54.8 & 18.9 & 33.0 & 25.9 & \textbf{49.4} & 14.4 & 40.3 & 25.1 & 37.6 & 19.3 & 41.1 & 25.2 & 39.7 \\
        Complexity-CoT                  & 28.6 & 61.9 & 20.0 & 34.1 & 28.1 & 44.8 & 17.0 & 40.6 & \textbf{23.7} & 34.3 & 21.8 & 50.4 & 29.5 & 40.1 \\       

        \hline
        \multicolumn{15}{|l|}{\cellcolor{mygray} \textit{Multi-step Chain-of-Thought}} \\
        \hline
        PS-CoT                          & 21.3 & 52.8 & 17.9 & 34.1 & 21.8 & 46.1 & 16.4 & \textbf{42.5} & 18.1 & 39.1 & 16.0 & 51.1 & \textbf{23.2} & 38.5 \\
        \textbf{Detective Thinking Prompt}   & \textbf{45.5} & \textbf{61.5} & \textbf{20.9} & \textbf{36.4} & 20.1 & 45.1 & \textbf{18.9} & 42.2 & 22.3 & \textbf{43.8} & \textbf{25.2} & \textbf{52.4} & 20.7 & 40.5 \\

        \hline
        \multicolumn{15}{|l|}{\cellcolor{mygray} \textit{Question with Extra Key Information}} \\
        \hline
        
        Naive w/ Key Info               & 65.4 & 64.8 & 42.9 & 34.9 & 48.3 & 58.1 & 22.7 & 47.9 & 47.1 & 44.5 & 48.7 & 47.6 & 61.3 & 48.9 \\
        Naive w/ Key Info (3-shot)      & 63.6 & 40.1 & 39.5 & 45.6 & 43.7 & 45.5 & 35.8 & 50.2 & 31.6 & 49.7 & 32.5 & 48.3 & 67.4 & 49.6 \\
        Naive w/ Answer                 & 47.3 & 99.0 & 20.3 & 94.5 & 36.5 & 98.0 & 23.0 & 57.0 & 18.0 & 69.4 & 17.9 & 47.9 & 13.7 & 56.9 \\
        Naive w/ Answer (3-shot)        & 55.3 & 77.6 & 18.3 & 82.5 & 35.1 & 97.0 & 20.8 & 49.6 & 16.3 & 71.3 & 14.9 & 35.5 & 14.9 & 61.1 \\
    \hline
    \end{tabular}
    }
    \caption{
    The performance of baseline models under renowned prompt engineering techniques is presented.
    Results in bold indicate the best results achieved without additional information.
    }
    \label{tab:prompt_experiments}
    \vspace{-5mm}
\end{table*}

\begin{table}[t]
\centering
\resizebox{0.9\columnwidth}{!}{
\begin{tabular}{|l|cc|c|c|}
\hline
 &
 \multicolumn{2}{c|}{Detective} & \multicolumn{1}{c|}{HotPotQA}  & \multicolumn{1}{c|}{Reclor} \\
&KeyInfo.&Acc.&RougeL-F.&Acc.\\
\hline

\multicolumn{5}{|l|}{\cellcolor{mygray} \textit{Llama2-base}} \\
\hline
Naive                           & 10.8 & 47.5 & 30.6 & 36.7 \\
\textbf{DT Prompt}              & 20.7 & 40.5 & 32.1 & 37.5 \\ 
\textbf{DT Prompt w/ MR Chat}   & 23.6 & 45.1 & 33.6 & 35.2 \\ 
\textbf{DT FT w/ Detective}     & \textbf{38.6} & \textbf{56.7} & \textbf{37.2} & \textbf{39.6} \\
\textbf{DT FT w/ Generated}     & 32.4 & 44.6 & 32.8 & 33.5 \\

\hline
\multicolumn{5}{|l|}{\cellcolor{mygray} \textit{Llama2-Chat}} \\
\hline
Naive                           & 10.8 & 47.5 & 36.3 & 38.8 \\
\textbf{DT Prompt}              & 25.2 & 52.4 & 39.7 & 42.6 \\ 
\textbf{DT Prompt w/ MR Chat}   & 22.7 & 50.1 & 37.1 & 40.5 \\ 
\textbf{DT FT w/ Detective}     & \textbf{40.9} & \textbf{58.3} & \textbf{41.7} & \textbf{45.5} \\
\textbf{DT FT w/ Generated}     & 34.6 & 50.5 & 38.6 & 37.1 \\ 

\hline
\multicolumn{5}{|l|}{\cellcolor{mygray} \textit{ChatGLM3-Base}} \\
\hline
Naive                           &  9.7 & 39.6 & 26.8 & 30.1 \\
\textbf{DT Prompt}              & 22.3 & 43.8 & 25.4 & 31.9 \\ 
\textbf{DT Prompt w/ MR Chat}   & 23.6 & 45.3 & 26.0 & 32.4 \\ 
\textbf{DT FT w/ Detective}     & \textbf{37.6} & \textbf{50.8} & \textbf{34.2} & \textbf{36.7} \\
\textbf{DT FT w/ Generated}     & 35.4 & 43.6 & 30.9 & 32.9 \\  

\hline
\multicolumn{5}{|l|}{\cellcolor{mygray} \textit{ChatGLM3-Chat}} \\
\hline
Naive                           & 15.3 & 41.3 & 31.8 & 33.0 \\
\textbf{DT Prompt}              & 18.9 & 42.2 & 37.6 & 38.9 \\ 
\textbf{DT Prompt w/ MR Chat}   & 14.6 & 41.9 & 35.4 & 38.4 \\ 
\textbf{DT FT w/ Detective}     & \textbf{27.1} & \textbf{56.3} & \textbf{42.3} & \textbf{41.7} \\
\textbf{DT FT w/ Generated}     & 24.6 & 43.5 & 38.5 & 39.1 \\ 

\hline
\end{tabular}
}
\caption{
A detailed comparison of baseline models' performances utilizing Self-Question Prompt and Fine-tuning methodologies is also provided.
Outcomes rendered in bold signify the most superior results within the same model under these experimental conditions.
}
\label{tab:DT_experiments}
\vspace{-5mm}
\end{table}

\begin{figure}[t]
    \centering
    \resizebox{0.8\columnwidth}{!}{
    \includegraphics{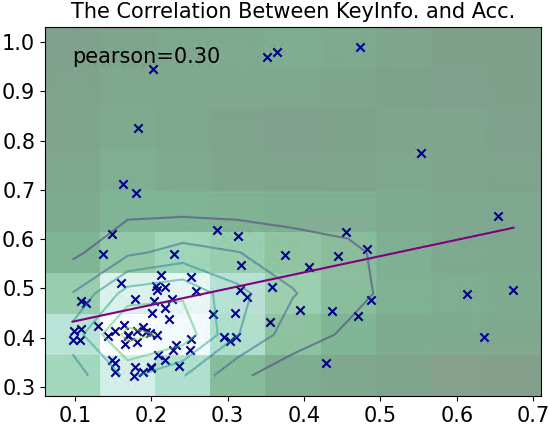}}
    \caption{The Pearson Correlation between the KeyInfo. metric and the Accuracy metric across all models and prompt methods.}
    \label{fig:perason}
\vspace{-7mm}
\end{figure}

\subsection{Overall Setup}

\textbf{Models:}
In order both to test the best performance of the LLMs and to ensure replicability, we have used a number of eminent models from both the API-based and the open source domains.
These include GPT4-turbo (GPT4)~\cite{openai2023gpt4}, GPT3.5-turbo (GPT35)~\cite{openaichatgpt}, Llama2-7b-Base (llama2-base), Llama2-7b-Chat (llama2-chat)~\cite{touvron2023llama}, GLM4 (GLM4)~\cite{zheng2023judging}, ChatGLM3-6b-Base (chatglm3-base), and ChatGLM3-6B-Chat (chatglm3-chat)~\cite{xu2023wizardlm}. The experimentation was conducted using the official APIs for GPT4-turbo, GPT-3.5-turbo, and GLM-4 between January 10 and January 29, 2024.

\textbf{Metrics:}
The DetectBench, which consists of multiple choice questions, uses Accuracy as a metric to assess the likelihood of model correctness in answer selection, which is denoted as \textbf{Acc}. 
In addition, the benchmark evaluates the ability of models to identify important information from context, a task similar to underline type machine reading comprehension.
Accuracy should be used to assess the model's ability to detect clues, but RougeL is used as a metric given the challenging nature of requiring LLM to generate content snippets directly from context, which denote as \textbf{KeyInfo}.

\subsection{Performance with Different Prompt Engineering}
\subsubsection{Experimental Setup}
\textbf{Baselines:}
A range of prompt engineering methods were analyzed for comparative insights:
\texttt{\textbf{Naive}}, which simply inputs ``Context'', ``Question'', and ``Options'' into LLMs for answers.
\texttt{\textbf{Self-CoT}}~\cite{kojima2022large}, applying a step-by-step reasoning prompt.
\texttt{\textbf{Auto-CoT}}~\cite{zhang2022automatic}, which automates Chain of Thought (CoT) demonstrations, evaluated in a three-shot setting due to its non-zero-shot design.
\texttt{\textbf{Self-Consistency}}~\cite{wang2022self}, summarizing multiple outputs from the same model to derive a final answer.
\texttt{\textbf{Complexity-CoT}}~\cite{fu2022complexity}, selecting the longest reasoning steps among all outputs.
\texttt{\textbf{Plan-and-Solve CoT (PS-CoT)}}, focusing on problem deconstruction before solution.
\texttt{\textbf{Detective Thinking Prompt}}~\cite{wang2023plan}, introduced in this study.
\texttt{\textbf{Naive /w Key Info}} and \texttt{\textbf{Naive /w Answer}}, enhancing inputs with ``Key Information'' and the ``Answer'' respectively.

Some methods are not included in the experiments:
Methods that involve a self-checking process, such as Tree of Thought~\cite{yao2023tot} and Graph of Thought~\cite{besta2023got}, were excluded because common sense reasoning is difficult to self-check during intermediate processes.
Methods such as Reflexion~\cite{shinn2023reflexion}, which increase the probability of a correct answer by injecting model error, were ruled out due to the priori information would be incured in the choice of options in a option-based QA setting.

\textbf{Demonstration:}
Demonstration incorporates correct answers in the test data format and a small number of examples to improve understanding of the output format and knowledge acquisition.
The Naive Prompt method appends answers after training data examples, while Auto-CoT guides the LLM to generate reasoning processes that are aligned with the ``Context'', ``Question'' and ``Answer''.

\subsubsection{Analysis}
Tab.~\ref{tab:prompt_experiments} displays the performance of all baseline models across different prompt methods.

\textbf{Varied Prompt Engineering Method Efficacy:}
Data shows that proprietary models like GPT4, GPT3.5, and GLM4 excel beyond open-source models such as ChatGLM3 and Llama2. Significant accuracy gains were observed with GPT4 and GLM4 using prompt engineering, whereas methods like Self-CoT saw a minor performance reduction in GPT3.5, ChatGLM3, and Llama2. This indicates that while advanced models benefit from prompt-guided reasoning, imposing such techniques on models with less sophisticated reasoning abilities may lead to performance decrements.

\textbf{Key Information Detection Shortcomings:}
A general shortfall in key information detection was noted, especially with GPT4-Turbo's average accuracy standing at 40\%. While accurate answers don't always require pinpointing key information, a direct correlation exists between identifying such information and answer accuracy. Directly presenting key information to models notably improved RougeL scores and answer accuracy, emphasizing the importance of precise key information identification.

\textbf{Reduced Demonstration Effectiveness:}
The historical utility of demonstrations in enhancing model response parsing has diminished as models have grown adept at interpreting complex instructions. Integration of three-shot demonstrations resulted in unstable performance across various prompt methods and model types~\cite{gu2023xiezhi}.

\textbf{Detective Thinking Prompt Superiority:}
The Detective Thinking method, unique to this study, markedly improved key information detection and reasoning across models. This approach not only enhanced accuracy but also demonstrated a broader efficacy compared to other prompt engineering strategies, reinforcing its value in augmenting model understanding and reasoning capabilities.

\begin{figure*}[!ht]
    \centering
    \resizebox{0.8\textwidth}{!}{
    \includegraphics{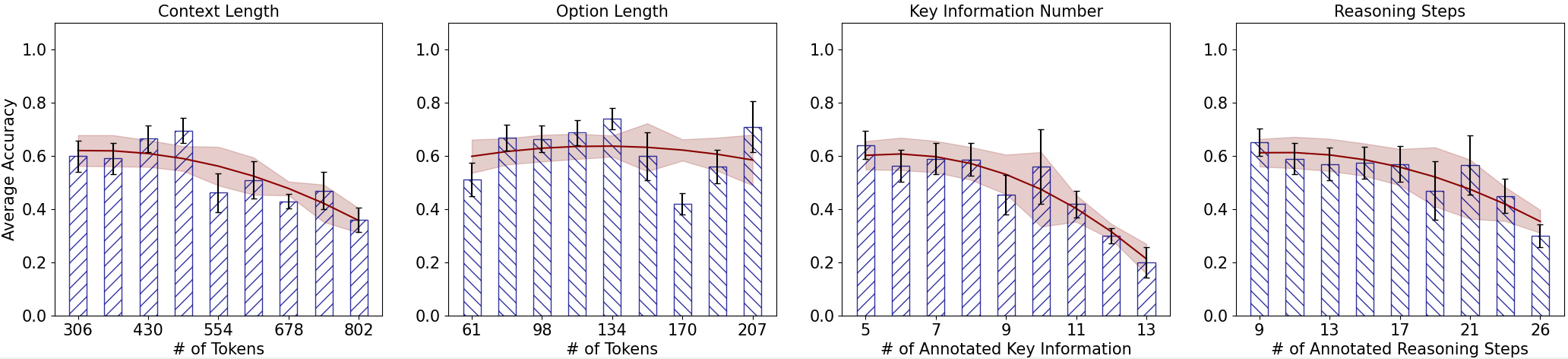}
    }
    \vspace{-2mm}
    \caption{The performance of GPT4-Turbo is correlated with the context length, option length, the quantity of clues, and the number of reasoning steps involved.}
    \label{fig:gpt4_performance}
\vspace{-5mm}
\end{figure*}

\begin{figure}[!ht]
    \centering
    \resizebox{0.7\columnwidth}{!}{
    \includegraphics{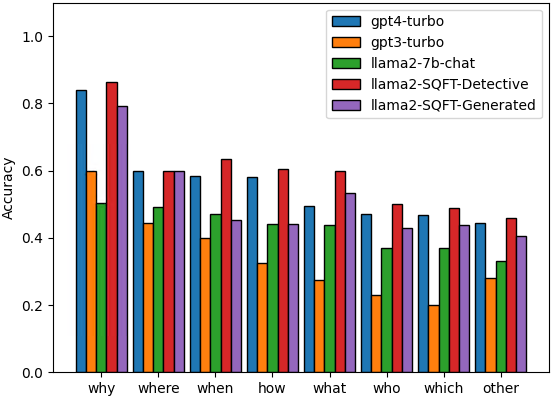}
    }
    \vspace{-2mm}
    \caption{The performance of various models varies across different Question Types.}
    \label{fig:performance_category}
\vspace{-5mm}
\end{figure}

\subsection{Optimizing Model Capabilities through Fine-Tuning}
\subsubsection{Experimental Setup}
\textbf{Baselines:}
We used four open source models to explore the role of fine-tuning in enhancing model performance.
Our focus is on evaluating the effectiveness of the Detective Thinking Prompt~(DT Prompt) applied directly, the DT Prompt within multi-round dialogues~(DT Prompt w/ MR Chat), fine-tuning using DetectBench data~(DT FT w/ Detective), and generating fine-tuning data from DetectBench context, question, and answer~(DT FT w/ Generated). 
A subset of 398 samples from the training dataset was used for fine-tuning over 3 epochs with the AdamW optimiser, as detailed in the Appendix~\ref{appendix_training}.
And Appendix~\ref{appendix_prompt} provides detailed descriptions of the prompts used in each method.




\subsubsection{Insights and Evaluations}

\textbf{Effects of Transitioning to Multi-Round Dialogues:}
Our analysis revealed that transitioning from single-round to multi-round dialogues negatively influences chat model accuracy by 1.3\%, while base models experience a 3.0\% accuracy enhancement. This differential impact suggests that base models, due to their heightened sensitivity to context, benefit from multi-round dialogues as a mechanism to distill relevant information. On the contrary, chat models, which are adept at key information extraction, face performance setbacks when dialogue is fragmented into multiple segments.

\textbf{Enhancements from DetectBench Data in Fine-Tuning:}
Utilizing DetectBench data for Detective Thinking Fine-tuning significantly boosts key information detection and reasoning skills in models. The observed post-fine-tuning improvements include a 15.2\% increase in key information detection accuracy and a 10.5\% uplift in overall model performance. These results underscore the DetectBench dataset's effectiveness in refining models' information processing and reasoning faculties.

\subsection{In-depth Performance Analysis}
\subsubsection{Performance Influencing Factors}
The analysis of GPT4-Turbo's performance, as detailed in Figure~\ref{fig:gpt4_performance}, highlights the effects of varying Context Length and Options Length on model accuracy. A notable decline in accuracy was observed as Context Length increased from 400 to 800 words, with accuracy dropping from approximately 65\% to 35\%. Additionally, the variability in Options Length indicated a struggle with reasoning complexity at both extremes of option length.

An examination of our annotations against model performance revealed a strong correlation between the volume of Key Information, reasoning depth, and performance metrics. Specifically, as the number of key information instances and reasoning depth escalated, a marked decrease in model accuracy was recorded, affirming the relationship between question complexity and model effectiveness.

\subsubsection{Varied Responses to Different Question Types}
The performance variation across different question types, as presented in Figure~\ref{fig:performance_category}, shows models excelling in answering ``Why'' and ``Where'' questions, with the fine-tuned Llama-2 model achieving an impressive 90\% accuracy. In contrast, the accuracy for ``Who'', ``Which'' and other question types hovered around 50\%. This disparity suggests that while models effectively handle questions requiring an understanding of processes and environments, they struggle with questions that demand sophisticated entity recognition and relational discernment, pinpointing areas for future model enhancement.

\section{Conclusion}

In this paper, we introduce the DetectBench, which integrates information retrieval and reasoning, catering to the current demand for task-oriented complex information retrieval.
This involves identifying key information from a plethora of data and conducting in-depth reasoning based on this key information to accomplish tasks.
Additionally, we propose a novel type of prompt engineering and fine-tuning method termed the Self-Question Framework, designed to concurrently augment model performance in key information detection and commonsense reasoning.

\section{Limitations}

The DetectBench is conceptualized to facilitate the assessment of machine learning models' capabilities in simultaneously detecting information and engaging in commonsense reasoning. However, when juxtaposed with the complexity and breadth of information encountered in real-world scenarios, the data encompassed within detective reasoning puzzles appears markedly condensed.

The implementation of a Detective Thinking Prompt has demonstrated efficacy in enhancing the performance of models on the DetectBench. Nevertheless, this strategy is predominantly effective for tasks necessitating the extraction and inference of pivotal information from extensive datasets. Its efficacy diminishes substantially in scenarios where the information at hand is minimal and necessitates the incorporation of implicit knowledge derived from common sense or experiential understanding.

\section{Ethical Concerns}

Given that a benchmark concentrating on detective deduction puzzles is predisposed to encompass a multitude of sensitive subjects, including but not limited to homicides and thefts. 
If not meticulously moderated, there exists a risk that models might refuse responding to sensitive questions for security purposes, consequently disadvantaging models that prioritize higher security standards.
Moreover, models that undergo fine-tuning using benchmark data may inadvertently amplify security vulnerabilities.
Considerable effort and resources have been allocated towards mitigating the ethical dilemmas associated with the Detective Reasoning Benchmark, with the dual objectives of ensuring that models committed to security do not eschew responding to sensitive questions and that the utilization of our dataset does not compromise model security.





\bibliography{anthology,custom}
\appendix
\section{Training Details}
\label{appendix_training}
For the models llama2-7b-base, llama2-7b-chat, ChatGPT3-6b-base, and ChatGPT3-6b-chat, we executed two distinct training methodologies:
\begin{enumerate}
    \item Directly utilizing the training data from the Detective Reasoning Benchmark to compose the Detective Thinking Finetune data.
    \item Employing the ``Context'', ``Question'', and ``Answer'' in Detective Reasoning Benchmark to automatically generate Detective Thinking Finetune data.
\end{enumerate}

The specific training parameters are detailed in Table \ref{tab:training_param}.

\begin{table*}[!ht]
    \centering
    \myred{
    \begin{tabular}{|c|c|c|c|c|}
    \hline
    \multicolumn{5}{|c|}{Training Detail} \\
    \hline
    \# of Samples & \# of Tokens & \# of epochs & warm\_up steps & learning rate \\
    396 & 162,868 & 3 & 200 & 1e-5  \\
    \hline
    \end{tabular}
    }
    \caption{\myred{All the parameter setting in the training process.}}
    \label{tab:training_param}
\end{table*}

\section{Detail about Manual Annotation}
\label{appendix_annotators}
\subsection{Details about Annotators}
The annotators for this research are the authors of this paper themselves, who are experts in the field of Computer Science and Cognitive Psychology.
The entire annotation process was under the stringent supervision and scrutiny of the first author of this paper.
\subsection{Annotation Tasks and Goals}
The purpose of the manual annotation tasks was twofold.
The first goal was to obtain comprehensive annotated datasets that encapsulate the essential features of the target text, which could be further leveraged for tasks such as training, testing, and model evaluation.
The second goal was to provide a detailed, rigorous, and systematic assessment of the annotated data quality to assess its fit and reliability for the subsequent analysis.
All the detailed annotation tasks and targets are listed in Tab.~\ref{tab:annotation_request}.
\subsection{Case of Annotation}
In our efforts to delineate the complex annotation process and ensure the replicable rigor of experiments, this section provides an in-depth display of the manual annotation cases.
The aim is to elucidate the categorical distinctions and precise definitions adopted in the annotations, thereby facilitating fellow researchers in ascertaining the veracity of the annotated data.
Representative cases from the annotation process have been cataloged in Tab.~\ref{tab:appendix_annotation_example} for comprehensive reference and understanding.

\begin{table*}[t]
    \centering
    \resizebox{\textwidth}{!}{
    \begin{tabular}{|p{0.3\textwidth}|p{0.69\textwidth}|}
    \hline
         \textbf{Task} & \textbf{Requirements}\\
         \hline
         \multirow{5}{*}{Question Verification}
         & 1.1 Delete if answering the question requires non-text information, like audio or image. \\
         & 1.2 Delete if there is a substantial amount of mathematical content or involve of too much domain knowledge. \\
         & 1.3 Delete if there is no ample presence of daily scenarios. \\
         & 1.4 Delete if the answer is not correct. \\
         & 1.5 Delete if there is any discrimination or bias concerning gender, race, nation, or religion. \\

         \hline
         \multirow{3}{*}{Question Rewrite} 
         &2.1 Standardize the Expression. \\
         &2.2 Rewrite a decent answer to the question.\\
         &2.3 Separate ``Question''and ``Context''.\\
         &2.4 Write decent and confusing ``Options'' of the question.\\
         
         \hline
         \multirow{3}{*}{Clue Graph Construction} 
         &3.1 Regenerate or rewrite if the ``Key Information of Context'' cannot exact match to the text in ``Context''.\\
         &3.2 Regenerate or rewrite if the connection or reasoning is redundant.\\
         &3.3 Delete the question or rewrite it there lack of important reasoning processes or connections in Clue Graph.\\
    \hline
    \end{tabular}}
    \caption{All tasks that require manual annotation, along with the specific requirements for each task.}
    \label{tab:annotation_request}
\end{table*}
\begin{table*}[t]
\centering
\resizebox{\textwidth}{!}{
\begin{tabular}{|c|l|l|}
\hline
Task &
  Requirements &
  Cases \\ \hline
\multirow{5}{*}{\begin{tabular}[c]{@{}c@{}}Question \\ Verification\end{tabular}}
&
\begin{tabular}[c]{@{}l@{}}Delete if answering the question \\ requires non-text information, like \\ audio or image.\end{tabular} 
&
\begin{tabular}[c]{@{}l@{}}
Context: ``Listen to the following music clip...''\\
Question: ``What instrument is playing?''\\ 
Hint: ``Consider the type of information required to answer the question.''\\ 
Answer: ``Piano''
\end{tabular} \\ \cline{2-3} 
 &
  \begin{tabular}[c]{@{}l@{}}Delete if there is a substantial \\ amount of mathematical content.\end{tabular} &
  \begin{tabular}[c]{@{}l@{}}Context: ``Consider the mathematical proof of Fermat's Last Theorem...''\\ Question: ``Can you explain the proof?''\\ Hint: ``Focus on the subject matter of the proof.''\\ Answer: ``It's a complex proof involving modular forms...''\end{tabular} \\ \cline{2-3} 
 &
  \begin{tabular}[c]{@{}l@{}}Delete if there is no ample presence \\ of daily scenarios.\end{tabular} &
  \begin{tabular}[c]{@{}l@{}}Context: ``In a quantum physics experiment...''\\ Question: ``What is the result?''\\ Hint: ``Consider the context of the experiment.''\\ Answer: ``A specific quantum state''\end{tabular} \\ \cline{2-3} 
 &
  Delete if the answer is not correct. &
  \begin{tabular}[c]{@{}l@{}}Context: ``The cat is on the roof''\\ Question: ``Where is the cat?''\\ Hint: ``Check the location mentioned in the context.''\\ Answer: ``In the garden''\end{tabular} \\ \cline{2-3} 
 &
  \begin{tabular}[c]{@{}l@{}}Delete if there is any discrimination \\ or bias concerning gender, race, \\ nation, or religion.\end{tabular} &
  \begin{tabular}[c]{@{}l@{}}Context: ``All people from X are lazy...''\\ Question: ``What are people from X like?''\\ Hint: ``Considering the description of X.''\\ Answer: ``Lazy''\end{tabular} \\ \hline
\multirow{4}{*}{\begin{tabular}[c]{@{}c@{}}Question \\ Rewrite\end{tabular}} &
  Standardize the Expression. &
  \begin{tabular}[c]{@{}l@{}}Original: ``$\langle$ /span $\rangle$ A family decides to move into the city and looks for a house. \textbackslash n \textbackslash n There are three ...'' \\
  Rewritten: ``A family decides to move into the city and looks for a house. There are three ... ''\end{tabular} \\ \cline{2-3} 
 &
  \begin{tabular}[c]{@{}l@{}}Rewrite a decent answer to the \\ question.\end{tabular} &
  \begin{tabular}[c]{@{}l@{}}Original Answer: ``This is a famous question, in my thought, the answer is ......''\\ Rewritten Answer: ``The answer is ......''\end{tabular} \\ \cline{2-3} 
 &
  Separate ``Question'' and ``Context''. &
  \begin{tabular}[c]{@{}l@{}}Original: \\  Context and Question: ``In 1862, during the American Civil War, the Battle \\         \quad                           of Antietam took place near Sharpsburg, Maryland... \\        \quad                              What was the significance of the Battle of Antietam?''\\ Separated:\\   Context: ``In 1862, during the American Civil War, the Battle of Antietam \\      \quad                      took place near Sharpsburg, Maryland...''\\   Question: ``What was the significance of the Battle of Antietam?''\end{tabular} \\ \cline{2-3} 
&
\begin{tabular}[c]{@{}l@{}}Write decent and confusing ``Options''\\ of the question.\end{tabular} &
\begin{tabular}[c]{@{}l@{}}
     Context: \\
     As the investigation unfolded, the police tape crisscrossed the snow-laden streets, casting eerie shadows under\\ \quad
     the moonlit night. The neighborhood, usually quiet and reclusive... \\
     Question: \\
     Do you think this young man is guilty or not? \\
     Answer: \\
     The young man could not have seen the murderer's detailed features due to the room's conditions \\
     Options: \\
     A) The young man was telling the truth, and the blond boyfriend was the murderer.\\
     B) The young man lied about the time of witnessing the murder to mislead the investigation.\\
     C) The young man could not have seen the murderer's detailed features due to the room's conditions.\\
     D) The victim had another visitor that night who was the real murderer\\
\end{tabular}\\ \hline

\multirow{3}{*}{\begin{tabular}[c]{@{}c@{}}Clue Graph\\Construction\end{tabular}} &
  \begin{tabular}[c]{@{}l@{}}Regenerate or rewrite if the ``Key\\ Information of Context'' cannot exact\\ match to the text in ``Context''.\end{tabular} &
  \begin{tabular}[c]{@{}l@{}}Original \\   Context: ``On a snowy winter night ...''\\   Key Information: ``On a blustery snowy winter night''\\ Rewritten\\   Key Information: ``On a snowy winter night ...''\end{tabular} \\ \cline{2-3} 
 &
  \begin{tabular}[c]{@{}l@{}}Regenerate or rewrite if the connection\\ or reasoning is redundant\end{tabular} &
  \begin{tabular}[c]{@{}l@{}}Original\\   Reasoning Process: ``Serene snowy setting + Murder at 68 King's West Road around 8pm \\ \quad→ Peaceful night disrupted by murder \\   Rewritten: \\Reasoning Process: \sout{``Serene snowy setting + Murder at 68 King's West Road around 8pm }\\ \quad \sout{→ Peaceful night disrupted by murder} \\\end{tabular} \\ \cline{2-3} 
 &
  \begin{tabular}[c]{@{}l@{}}Delete the question or rewrite it there\\ lack of important reasoning processes\\ or connections in Clue Graph.\end{tabular} &
  \begin{tabular}[c]{@{}l@{}}-\end{tabular} \\ \hline
  
\end{tabular}}
\caption{The examples in our annotation process}
\label{tab:appendix_annotation_example}
\end{table*}

         

\section{Experiments Details}
\label{appendix_experiment}

\subsection{Parameters in Inference}
Our experiments involved two types of hyperparameters.
The first type pertains to the seeds of random numbers used in various Python libraries, while the second type refers to the hyperparameters used when invoking the AutoCausalLM class from the transformers library for generation.
We configured our settings as demonstrated in Table~\ref{tab:hyperparameter}.

\begin{table*}[!ht]
    \centering
    \resizebox{\textwidth}{!}{
    \myred{
    \begin{tabular}{|c|c|c|c|c|}
    \hline
    \multicolumn{5}{|c|}{Random Seed} \\
    \hline
    torch.manual\_seed & torch.cuda.manual\_seed\_all & numpy.random.seed & random.seed & torch.backends.cudnn.deterministirc \\
    42 & 42 & 42 & 42 & True \\
    \hline
    \hline
    \multicolumn{5}{|c|}{AutoCausalLM} \\
    \hline
    temperature & top\_p & top\_k & num\_beams & max\_new\_token \\
    0.95 & 0.95 & 5 & 2 & 2000  \\
    \hline
    \end{tabular}
    }}
    \caption{\myred{All the parameter setting in model inference in our experiments.}}
    \label{tab:hyperparameter}
\end{table*}

\subsection{Prompt Details}
\label{appendix_prompt}
This section primarily showcases the prompts employed by all Prompt Engineers throughout the experiment.

Table~\ref{tab:naive} displays the \texttt{Naive} prompts,
Table~\ref{tab:naivekeyinfo} presents the \texttt{Naive w/ Key Info} prompts,
Table~\ref{tab:naiveanswer} outlines the \texttt{Naive w/ Answer} prompts,
Table~\ref{tab:selfcot} features the \texttt{Self-CoT} prompts,
Table~\ref{tab:selfconsistency} exhibits the \texttt{Self-Consistency} prompts,
Table~\ref{tab:complexitycot} reveals the \texttt{Complexity-CoT} prompts,
Table~\ref{tab:pscot} shows the \texttt{PS-CoT} prompts,
Table~\ref{tab:selfquestion} displays the \texttt{Detective Thinking Prompt} prompts, and




\begin{table*}[!ht]
    \centering
    \resizebox{\textwidth}{!}{
    \begin{tabular}{|p{\textwidth}|}
    \hline
\# -*- coding: utf-8 -*-\\

Variables:\\
!<INPUT 0>! -- Context\\
!<INPUT 1>! -- Question\\
!<INPUT 2>! -- Options\\

<commentblockmarker>\#\#\#</commentblockmarker>

Below I will give you a detective reasoning question, please summarize the key clues in this question based on the Context, the options and choose the answer you think is correct.
Note: When generating the answer, please only output the serial number of the option.

\#\#\# Context:\\
!<INPUT 0>!

\#\#\# Question:\\
!<INPUT 1>!

\#\#\# Options:\\
!<INPUT 2>!

Your output will contain the following:
\#\#\# Key Information: Please output what you consider to be the key information in the Context. Please note that the key information needs to be directly from the Context, i.e. it is a string originally in the Context that can be matched directly to the original text by string matching.
\#\#\# Answer: please output only the serial numbers.

Please follow the format below for your output:

\#\#\# Key Information:
xxxxx\\

\#\#\# Answer:
1/2/3/4\\
\bottomrule
\end{tabular}}
\caption{Prompt of \texttt{Naive} method}
\label{tab:naive}
\end{table*}

\begin{table*}[!ht]
    \centering
    \resizebox{\textwidth}{!}{
    \begin{tabular}{|p{\textwidth}|}
    \hline
\# -*- coding: utf-8 -*-\\

Variables:\\
!<INPUT 0>! -- Context\\
!<INPUT 1>! -- Question\\
!<INPUT 2>! -- Key Information
!<INPUT 3>! -- Options\\

<commentblockmarker>\#\#\#</commentblockmarker>

Below I will give you a detective reasoning question, please summarize the key clues in the question based on the Context, the options, and the answer, and choose the answer you think is correct.
Note: When generating the answer, please output only the serial number of the option.\\
\\
\#\#\# Context:\\
!<INPUT 0>!\\
\\
\#\#\# Question:\\
!<INPUT 1>!\\
\\
\#\#\# Key Information:\\
!<INPUT 2>!\\
\\
\#\#\# Option:\\
!<INPUT 3>!
\\
Your output will contain the following:\\
\#\#\# Key Information: Please output what you consider to be the key information in the Context. Please note that the key information needs to be directly from the Context, i.e. it is a string originally in the Context that can be matched directly to the original text by string matching.\\
\#\#\# Answer: please output only the serial numbers.\\
\\
Please follow the format below for your output:\\
\\
\#\#\# Key Information:\\
xxxxx\\
\\
\#\#\# Answer:\\
1/2/3/4\\
    \hline
    \end{tabular}}
    \caption{Prompt of \texttt{Naive w/ Key Information} method}
    \label{tab:naivekeyinfo}
\end{table*}

\begin{table*}[!ht]
    \centering
    \resizebox{\textwidth}{!}{
    \begin{tabular}{|p{\textwidth}|}
    \hline
\# -*- coding: utf-8 -*-\\

Variables:\\
!<INPUT 0>! -- Context\\
!<INPUT 1>! -- Question\\
!<INPUT 2>! -- Options\\
!<INPUT 3>! -- Answer\\
\\
<commentblockmarker>\#\#\#</commentblockmarker>\\
\\
Below I will give you a detective reasoning question, please summarize the key clues in the question based on the Context, the options, and the answer, and choose the answer you think is correct.\\
Note: When generating the answer, please output only the serial number of the option.\\
\\
\#\#\# Context:\\
!<INPUT 0>!\\
\\
\#\#\# Question:\\
!<INPUT 1>!\\
\\
\#\#\# Options:\\
!<INPUT 2>!\\
\\
\#\#\# Answer:
!<INPUT 3>!\\
\\
Your output will contain the following:\\
\#\#\# Key Information: Please output what you consider to be the key information in the Context. Please note that the key information needs to be directly from the Context, i.e. it is a string originally in the Context that can be matched directly to the original text by string matching.\\
\#\#\# Answer: please output only the serial numbers.\\
\\
Please follow the format below for your output:\\
\\
\#\#\# Key Information:
xxxxx\\

\#\#\# Answer:\\
1/2/3/4\\
    \hline
    \end{tabular}}
    \caption{Prompt of \texttt{Naive w/ Answer} method}
    \label{tab:naiveanswer}
\end{table*}

\begin{table*}[!ht]
    \centering
    \resizebox{\textwidth}{!}{
    \begin{tabular}{|p{\textwidth}|}
    \hline
\# -*- coding: utf-8 -*-\\
\\
Variables:\\
!<INPUT 0>! -- Context\\
!<INPUT 1>! -- Question\\
!<INPUT 2>! -- Options\\
\\
<commentblockmarker>\#\#\#</commentblockmarker>\\
\\
Below I will give you a detective reasoning question, please generate your thought process step by step based on the Context and the options and choose the answer you think is correct.\\
Note: When generating the answer, please output only the serial number of the option.\\
\\
\#\#\# Context:\\
!<INPUT 0>!\\
\\
\#\#\# Question:\\
!<INPUT 1>!\\
\\
\#\#\# Options:\\
!<INPUT 2>!\\
\\
Your output will contain the following:\\
\#\#\# Thought: please output your thinking process step by step.\\
\#\#\# Key Information: Please output what you think is the Key Information in the Context. Please note that the Key Information needs to be directly from the Context, i.e. it is a string originally in the Context that can be matched directly to the original text by string matching.\\
\#\#\# Answer: please output only the serial numbers.\\
\\
Please have your output follow the format below:\\
\\
\#\#\# Thought:\\
xxxxxx\\
\\
\#\#\# Key Information:\\
xxxxx\\
\\
\#\#\# Answers:\\
1/2/3/4\\
    \hline
    \end{tabular}}
    \caption{Prompt of \texttt{Self-CoT} method}
    \label{tab:selfcot}
\end{table*}

\begin{table*}[!ht]
    \centering
    \resizebox{\textwidth}{!}{
    \begin{tabular}{|p{\textwidth}|}
    \hline
\# -*- coding: utf-8 -*-\\
\\
Variables:\\
!<INPUT 0>! -- Demonstration\\
!<INPUT 1>! -- Context\\
!<INPUT 2>! -- Question\\
!<INPUT 3>! -- Options\\
\\
<commentblockmarker>\#\#\#</commentblockmarker>\\
\\
\#\#\# Demonstration\\
!<INPUT 0>!\\
\\
\#\#\# Context:\\
!<INPUT 1>!\\
\\
\#\#\# Question:\\
!<INPUT 2>!\\
\\
\#\#\# Options:\\
!<INPUT 3>!\\
\\
Your output will contain the following:\\
\#\#\# Thought: please output your thinking process step by step.\\
\#\#\# Key Information: Please output what you think is the key information in the topic. Please note that the key information needs to be directly from the question, i.e. it is the original string in the question, which can be matched directly to the original text by string matching.\\
\#\#\# Answer: When generating answers, please output only the serial numbers of the options.\\
\\
Please follow the format below for your output:\\
\\
\#\#\# Thought:\\
xxxxx\\
\\
\#\#\# Key Information:\\
xxxxx\\
\\
\#\#\# Answer:\\
1/2/3/4\\
    \hline
    \end{tabular}}
    \caption{Prompt of \texttt{Auto-CoT} method}
    \label{tab:auto_cot}
    
\end{table*}\begin{table*}[!ht]
    \centering
    \resizebox{\textwidth}{!}{
    \begin{tabular}{|p{\textwidth}|}
    \hline
\# -*- coding: utf-8 -*-\\
\\
Variables:\\
!<INPUT 0>! -- Context\\
!<INPUT 1>! -- Question\\
!<INPUT 2>! -- Options\\
\\
<commentblockmarker>\#\#\#</commentblockmarker>\\
\\
Below I will give you a detective reasoning question, please generate your thought process step by step based on the Context and the options and choose the answer you think is correct.\\
Note: When generating the answer, please output only the serial number of the option.\\
\\
\#\#\# Context:\\
!<INPUT 0>!\\
\\
\#\#\# Question:\\
!<INPUT 1>!\\
\\
\#\#\# Options:\\
!<INPUT 2>!\\
\\
Your output will contain the following:\\
\#\#\# Thought: please generate 5 completely different perspectives of your reflections based on the questions and options.\\
\#\#\# Summary: Please output a summary of all your thinking.\\
\#\#\# Key Information: Please output what you think is the Key Information in the Context. Please note that the Key Information needs to be directly from the Context, i.e. it is the original string in the Context, which can be matched directly to the original text by string matching.\\
\#\#\# Answer: please output only the serial numbers.\\
\\
Please have your output follow the format below:\\
\\
\#\#\# Thought:\\
1. xxxxxx\\
2. xxxxxx\\
3. xxxxxx\\
4. xxxxxx\\
5. xxxxxx\\
\\
\#\#\# Summarize:\\
xxxxxx\\
\\
\#\#\# Key Information:\\
xxxxx\\
\\
\#\#\# Answers:\\
1/2/3/4\\
    \hline
    \end{tabular}}
    \caption{Prompt of \texttt{Self Consistency} method}
    \label{tab:selfconsistency}
\end{table*}

\begin{table*}[!ht]
    \centering
    \resizebox{\textwidth}{!}{
    \begin{tabular}{|p{\textwidth}|}
    \hline
\# -*- coding: utf-8 -*-\\
\\
Variables:\\
!<INPUT 0>! -- Context\\
!<INPUT 1>! -- Question\\
!<INPUT 2>! -- Options\\
!<INPUT 3>! -- Longest Chain of Thought\\
\\
<commentblockmarker>\#\#\#</commentblockmarker>\\
\\
Below I will give you a detective reasoning question, please generate your thought process step by step based on the question and the options and choose the answer you think is correct.\\
Note: When generating the answer, please output only the serial number of the option.\\
\\
\#\#\# Context:\\
!<INPUT 0>!\\
\\
\#\#\# Question:\\
!<INPUT 1>!\\
\\
\#\#\# Options:\\
!<INPUT 2>!\\
\\
\#\#\# Chain of thought:\\
!<INPUT 3>!\\
\\
Your output will contain the following:
\#\#\# Key Information: Please output what you consider to be the key information in the topic. Please note that the key information needs to be directly from the topic, i.e. it is a string originally in the topic that can be matched directly to the original text by string matching.\\
\#\#\# Answer: please output only the serial numbers.\\
\\
Please follow the format below for your output:\\
\\
\#\#\# Key Information:\\
xxxxx\\
\\
\#\#\# Answer:\\
1/2/3/4\\
    \hline
    \end{tabular}}
    \caption{Prompt of \texttt{Complexity CoT} method}
    \label{tab:complexitycot}
\end{table*}

\begin{table*}[!ht]
    \centering
    \resizebox{\textwidth}{!}{
    \begin{tabular}{|p{\textwidth}|}
    \hline
\# -*- coding: utf-8 -*-\\

Variables:\\
!<INPUT 0>! -- Context\\
!<INPUT 1>! -- Question\\
!<INPUT 2>! -- Options\\
\\
<commentblockmarker>\#\#\#</commentblockmarker>\\
\\
Below I will give you a detective reasoning question, please generate your thought process step by step based on the Context and the options and choose the answer you think is correct.\\
Note: When generating the answer, please output only the serial number of the option.\\
\\
\#\#\# Context:\\
!<INPUT 0>!\\
\\
\#\#\# Question:\\
!<INPUT 1>!\\
\\
\#\#\# Options:\\
!<INPUT 2>!\\
\\
Your output will contain the following:\\
\#\#\# Thought: Please start with a general plan of how you intend to deal with the problem, and then think step-by-step about how to solve it based on your plan.\\
\#\#\# Key Information: please output what you think is the key information in the Context. Please note that the Key Information needs to be directly from the Context, i.e. it is the original string in the Context, which can be matched directly to the original text by string matching.\\
\#\#\# Answer: please output only the serial numbers.\\
\\
Please have your output follow the format below:\\
\\
\#\#\# Thought:\\
xxxxxx\\
\\
\#\#\# Key Information:\\
xxxxx\\
\\
\#\#\# Answer:\\
1/2/3/4\\
    \hline
    \end{tabular}}
    \caption{Prompt of \texttt{Plan and Solve CoT} method}
    \label{tab:pscot}
\end{table*}

\begin{table*}[!ht]
    \centering
    \resizebox{\textwidth}{!}{
    \begin{tabular}{|p{\textwidth}|}
    \hline
\# -*- coding: utf-8 -*-\\
\\
Variables:\\
!<INPUT 0>! -- Context\\
!<INPUT 1>! -- Question\\
!<INPUT 2>! -- Options\\
\\
<commentblockmarker>\#\#\#</commentblockmarker>\\
\\
Below I will give you a detective reasoning question, please generate your thought process step by step based on the Context and the options and choose the answer you think is correct.\\
Note: When generating the answer, please output only the serial number of the option.\\
\\
\#\#\# Context:\\
! <INPUT 0>!\\
\\
\#\#\# Question:\\
! <INPUT 1>!\\
\\
\#\#\# Options:\\
! <INPUT 2>!\\
\\
Your output will contain the following:\\
\#\#\# Clues: Feel free to summarize all possible clues in the Context\\
\#\#\# Connection: Feel free to correlate the clues you summarized above and introduce new clues that may exist.\\
\#\#\# Thought: Feel free to reason and think deeply about the clues you have summarized in the two steps above.\\
\#\#\# Summarize: Summarize all the thinking from the perspective of solving the problem in the Context.\\
\#\#\# Key Information: Please output what you think is the key information in the Context. Please note that the Key Information needs to be the direct content of the Context, i.e. it is the original string in the Context, which can be matched directly to the original text by string matching.\\
\#\#\# Answer: Please output only the serial number.\\
\\
Please have your output follow the format below:\\
\\
\#\#\# Clues:\\
xxxxxx\\
\\
\#\#\# Connection:\\
xxxxxx\\
\\
\#\#\# Thought:\\
xxxxxx\\
\\
\#\#\# Summarize:\\
xxxxxx\\
\\
\#\#\# Key Information:\\
xxxxx\\
\\
\#\#\# Answer:\\
1/2/3/4\\
    \hline
    \end{tabular}}
    \caption{Prompt of \texttt{Detective Thinking} method}
    \label{tab:selfquestion}
\end{table*}

\end{CJK}
\end{document}